\documentclass[11pt,a4paper]{article}
\usepackage[hyperref]{acl2021}
\usepackage{times}
\usepackage{latexsym}
\usepackage{amsmath}
\usepackage{graphicx}
\usepackage{makecell}
\usepackage{adjustbox}
\usepackage{multirow}
\usepackage{tabularx}
\usepackage{comment}
\usepackage{pbox}
\graphicspath{ {./} }

\usepackage{microtype}

\aclfinalcopy 

\title{Improving Factual Consistency of Abstractive Summarization on Customer Feedback}

\author{Yang Liu, Yifei Sun, Vincent Gao \\
   Amazon, Inc.  \\
  \texttt{\{yngliun, sunyifei, vincegao\}@amazon.com}}

\date{}

\begin{document}
\maketitle
\begin{abstract}
E-commerce stores collect customer feedback to let sellers learn about customer concerns and enhance customer order experience. Because customer feedback often contains redundant information, a concise summary of the feedback can be generated to help sellers better understand the issues causing customer dissatisfaction. Previous state-of-the-art abstractive text summarization models make two major types of factual errors when producing summaries from customer feedback, which are \textit{wrong entity detection} (WED) and \textit{incorrect product-defect description} (IPD). In this work, we introduce a set of methods to enhance the factual consistency of abstractive summarization on customer feedback. We augment the training data with artificially corrupted summaries, and use them as counterparts of the target summaries. We add a contrastive loss term into the training objective so that the model learns to avoid certain factual errors. Evaluation results show that a large portion of WED and IPD errors are alleviated for BART and T5. Furthermore, our approaches do not depend on the structure of the summarization model and thus are generalizable to any abstractive summarization systems.
\end{abstract}

\section{Introduction}
In order to improve customer order experience, most e-commerce stores allow customers to submit reviews or feedback via their post-order communication channels. Such customer feedback, usually in the form of short paragraphs of free texts, contains information reflecting the issues that customers experienced in their purchases. This information can be shared with sellers to bring their awareness on the problems in their products. However, customer feedback often include other contents that are irrelevant to the product issues. Such redundant information requires extra efforts for sellers to fully understand the customers major concerns, and sometimes even causes confusion. 

\begin{table}
\centering
\small
\begin{tabularx}{\columnwidth}{X}
\hline
\textbf{Source}: (...) I ordered this mouse for my new laptop. However, when I received it, I could see many scratches on the product. It looks like it has been used before.  (…) \\ 
\textbf{Reference Summary}: The \textcolor{blue}{mouse} delivered has many scratches. It looks like it has been used. \\ 
\textbf{Model Summary}: The \textcolor{red}{laptop} came with many scratches, looks like it has been used. \\
\hline
\textbf{Source}: (…) I checked the serial number and found it doesn't match the one on the website. This phone is not defective. I question the source of this product (…) \\ 
\textbf{Reference Summary}: The phone serial number doesn’t match the one on the website but the phone is \textcolor{blue}{not defective}. \\ 
\textbf{Model Summary}: This phone is \textcolor{red}{defective} and the serial number doesn’t match the one on the website. \\
\hline
\end{tabularx}
\caption{\label{table:error-examples} Examples of the two major factual errors: WED (upper) and IPD (lower). }
\end{table}

To reduce the redundancy, a concise summary of customer feedback can be provided where the information is concentrated on the product issues while other irrelevant contents are filtered out. Such summary allows sellers to quickly capture and comprehend the problems, and thus they can address buyer dissatisfaction more efficiently.

The problem of generating summaries from customer feedback is modeled as a text summarization task \citep{nallapati2016abstractive, allahyari2017text, gao2020standard} in the natural language processing (NLP) domain. Abstractive summarization models with transformer-based architecture have achieved success in a variety of summarization tasks \citep{lewis2020bart, raffel2020exploring, zhang2020pegasus, bao2020unilmv2}. Hence, we harnessed the recent state-of-the-art (SOTA) abstractive summarization models, BART \citep{lewis2020bart} and T5 \citep{raffel2020exploring}, and fine tuned the models for our specific summarization task. We aim to utilize summarization models to produce the summary that can correctly describe the product issues presented in customer feedback. However, from human evaluation results, we observed that the summary generated by these abstractive summarization models sometimes contains the information that is inconsistent with facts in the input text. Such factual inconsistencies have also been observed in previous studies \citep{cao2018faithful, kryscinski2019neural, kryscinski2020evaluating}. More specifically, we analyzed 75 inconsistent summaries obtained from human evaluations on more than 600 model-generated summaries. We found that around 70\% factual inconsistent summaries \footnote{The rest of the unfaithful summaries are due to miscellaneous factual errors that are hard to cluster.} follow two error patterns: \textit{wrong entity detection} (WED) and \textit{incorrect product-defect description} (IPD). The error of WED often occurs in the cases where the feedback text involves multiple entities but the models fail to detect the primary entity. For IPD, the generated summary contains the product-defect description that contradicts with the original description in the customer feedback. Table \ref{table:error-examples} shows the examples \footnote{Due to confidentiality, all customer feedback examples in this paper are composed by the authors.} of the two types of factual errors. 

In this work, we propose a set of methods in order to improve the factual consistency of abstractive summarization on customer feedback. We first introduce specific factual errors into each target summary to generate their negative counterpart. We then use such pair of consistent and inconsistent summaries with a contrastive loss term added in the training objective to enhance the model’s robustness against the two major factual errors. 

Our contributions are two folds. First, The proposed approaches with corrupted summary generation and contrastive loss augmentation do not pose requirements on the achitecture of the summarization model. Thus, they can be applied to any abstraction-based summarization model to improve the model faithfulness. Second, we test the proposed approaches on SOTA summarization algorithms such as BART and T5. Our approaches show large benefits in reducing the common factual errors in customer-feedback summarization.

\section{Related Work}
There have been increasing research attentions on improving the factual consistency of abstractive summarization models. Lots of priors work focused on different ways of adding external signals or constraints to enhance the summary generation. \citet{cao2018faithful} built a dual-attention framework so that the summary generation is conditioned on both the source document and extracted key information. \citet{li2018ensure} incorporated the entailment knowledge by utilizing entailment-aware encoder and decoder. With using the textual entailment, \citet{falke2019ranking} re-ranked the candidates summaries to select the summary that's better aligned with the source document. \citet{dou2020gsum} studied different external signals, including key sentences, keywords and relations, and used them in addition to the input text to guide the summary generation. \citet{mao2020constrained} constrained certain tokens to require them to be present in the summary. Similarly, \citet{yuan2020faithfulness} add constraints on the model to include certain attribute words in the product summarization. \citet{zhu2021enhancing} integrated information extraction and graph attention network into transformer-based seq2seq framework.

To identify and correct the unfaithful summaries, 
\citet{wang2020asking} proposed to use a question answering framework to check the faithfulness of the summary while \citet{dong2020multi} built a factual correction model that leverages knowledge learned from question answering models. \citet{kryscinski2020evaluating} trained a BERT-based model to classify whether the summary is factual consistent. \citet{cao2020factual} and \citet{zhu2021enhancing} developed factual corrector based on BART \citep{lewis2020bart} and UniLM \citep{dong2019unified}, as a post-processor to rectify factual errors from the upstream summarization model. They corrupted the reference summaries with artificial errors and used them as the negative samples for training the correctors. In our work, we also generate corrupted summaries as the negative counterparts of the target summaries. The difference is that, instead of building a separate corrector model, we directly engineer the training objective of the summarization model. By leveraging contrastive learning \cite{schroff2015facenet, khosla2020supervised}, we define contrasive losses to guide the output summary away from certain factual errors. 

\section{Proposed Approaches}
Our error analysis of customer-feedback summarizaton showed that most of the factual errors belong to two error types: WED and IPD. Hence, in our proposed approaches, we first apply rule-based transformations and introduce synthetic factual errors of the two error patterns into the target summaries. We then modify the training objective by adding the contrastive loss so as to guide the model to avoid those mistakes.

\subsection{Synthetic Factual Errors}
We augment the training data by applying two types of corruption methods on the target summary. The corruptions are designed to mimic the factual errors we observed. In the first method, we replace the named entities in the target summary with the other random entities of the same type in the source document. If no such replacement entity can be found in the source document, we randomly pick one from the top 50 appeared entities in our dataset. We used Spacy toolkit \cite{spacy} for the named entity extraction. In the second method, we use predefined rules to transform the product-defect description in the target summary. We detect the adjectives describing the product defect and switch their sentiment. There are two ways that we change the description. One is by adding negation word \textit{not} before the adjective. For example, we alter \textit{"product is broken"} to \textit{"product is not broken"}. If word \textit{not} is already presented, we will remove it instead. The other way is by switching a descriptive word to the one with opposite meaning, such as changing \textit{"opened"} to \textit{"sealed"}. Table \ref{table:corruption-examples} shows some examples of the corrupted summaries.

\begin{table}
\centering
\small
\begin{tabularx}{\columnwidth}{X}
\hline
\textbf{Source}: (...) I've bought cheese from this store for many times, and they were very good. So I think other products must be good too. Then I ordered several bottles of milk. But they are clearly expired (…) \\ 
\textbf{Reference Summary}: \textcolor{blue}{Milk} delivered is expired. \\ 
\textbf{Corrupted Summary}: \textcolor{red}{Cheese} delivered is expired. \\
\hline
\textbf{Source}: (…) The eggs I purchased have bad smells. They don't look like fresh eggs. (…) \\ 
\textbf{Reference Summary}: Eggs have \textcolor{blue}{bad} smells, and don't look like fresh eggs. \\ 
\textbf{Corrupted Summary}: Eggs have \textcolor{red}{good} smells, and don't look like fresh eggs.\\
\hline
\end{tabularx}
\caption{\label{table:corruption-examples} Examples of corrupted summaries. We replace the primary entity in the first example and switch the description in the second example.}
\end{table}

\subsection{Training Objective}
For each training sample, we now have a triplet $(d, s_+, s_-)$ consisting of the source document $d$, target summary $s_+$, and corrupted summary $s_-$. The summarization model takes $d$ as the input and generates the output $o$. Our training objective is to drive the model output $o$ to resemble $s_+$ while at the same time avoiding the factual errors presented in $s_-$. Inspired by contrastive learning \cite{schroff2015facenet, khosla2020supervised}, we compare different contrastive loss functions for model training.

\paragraph{Direct Contrast} Compared to the ordinary loss function for summarization, we add an extra term that takes into account the informration from corrupted summary:
$$\mathcal{L}_{DC} = \mathcal{L}(s_+, o) - \alpha * \mathcal{L}(s_-, o)$$
where $\mathcal{L}(s_+, o)$ is the cross entropy loss between $s_+$ and $o$, $\mathcal{L}(s_-, o)$ is the cross entropy loss between $s_-$ and $o$, and $\alpha$ is a tunable hyperparameter controlling the impact from the second term. The loss function will purely focus on the difference between $s_+$ and $s_-$ if $\alpha = 1.0$. Thus, we generally use small value for $\alpha$ to ensure the model will produce fluent summary.

\paragraph{Constrained Negative} Here, we add a margin term $M$ to constrain the value of $\mathcal{L}(s_-, o)$:
$$\mathcal{L}_{CN} = \mathcal{L}(s_+, o) + \alpha * \max{\Big( M - \mathcal{L}(s_-, o), 0 \Big)}$$
For easy negatives with $\mathcal{L}(s_-, o)>M$, their effects won't be taken into account during training as the model can confidently distinguish them from positive samples. 

\paragraph{Constrained Contrast} We augment the ordinary loss function for summarization with a constrained contrastive term:
\begin{equation*}
\begin{split}
\mathcal{L}_{CC} & = \mathcal{L}(s_+, o) + \\
& \alpha * \max{\Big( \mathcal{L}(s_+, o) + M - \mathcal{L}(s_-, o), 0 \Big) }
\end{split}
\end{equation*}
In this formula, the model is not only trained towards predicting correct labels but also deviating from certain factual errors extracted from the contrast between the negative and positive samples. 

\section{Experiments}
\subsection{Dataset}
We collected 10,000 samples of negative customer feedback from the post-order communication channels of e-commerce stores. We asked subject matter experts to generate summary for each customer feedback text with emphasis on extracting the information related to product issues. The summary is required to contain the (1) primary item names and (2) descriptions about the product defects associated with the items, if they are presented in the customer feedback. We use the human-produced summary as the target summary in model training. The train/test split ratio is 85:15.

\begin{table*}
\centering
\begin{tabular}{
>{\raggedright\arraybackslash}p{3.7cm}
>{\centering\arraybackslash}p{2cm}
>{\centering\arraybackslash}p{2cm}
>{\centering\arraybackslash}p{2cm}
}
\hline
\textbf{Model} & \textbf{ROUGE-1} & \textbf{ROUGE-2} & \textbf{ROUGE-L} \\
\hline
BART$_{+ corruption, \, {\mathcal{L}_{DC}}}$ & +0.30 & +0.36 & +0.49 \\
BART$_{+ corruption, \, {\mathcal{L}_{CN}}}$ & +0.54 & +0.01 & +0.59 \\
BART$_{+ corruption, \, {\mathcal{L}_{CC}}}$ & \textbf{+0.83} & \textbf{+1.12} & \textbf{+0.68} \\
\hline
\hline
T5$_{+ corruption, \, {\mathcal{L}_{DC}}}$ & +0.05 & -0.19 & +0.04 \\
T5$_{+ corruption, \, {\mathcal{L}_{CN}}}$ & +0.20 & +0.08 & +0.25 \\
T5$_{+ corruption, \, {\mathcal{L}_{CC}}}$ & \textbf{+0.45} & \textbf{+0.71} & \textbf{+0.43}\\
\hline
\end{tabular}
\caption{\label{table:ROUGE} Impact of our approaches on ROUGE scores. The reported numbers are relative changes of ROUGE scores compared to the ordinary fine-tuned BART and T5 models, respectively$^4$.}
\end{table*}

\subsection{Model}
We use two recently proposed abstractive summarization models, BART \citep{lewis2020bart} and T5 \cite{raffel2020exploring}, for customer-feedback summarization. We adopt the pretrained models from the HuggingFace implementation \footnote{\url{https://huggingface.co/transformers/}} and fine tune the models on our training dataset. Both models share the same training parameters including learning rate as 5e-5, $\alpha=0.05$ in $\mathcal{L}_{DC}$, $(\alpha=0.5, M=2.0)$ in $\mathcal{L}_{CN}$, and $(\alpha=0.5, M=5.0)$ in $\mathcal{L}_{CC}$.

\subsection{Evaluation metrics}
We employ the ROUGE-1, ROUGE-2, and ROUGE-L scores \citep{lin2004rouge} to ensure that our proposed methods do not degrade the fluency and continuity of the generated summary. These ROUGE scores measure the accuracy based on unigrams, bigrams, and longest subsequences.

We rely on the human evaluation to examine the factual consistency of the model output. We ask human annotators to classify the faithfulness of generated summary into \textit{consistent} and \textit{inconsistent} based on whether there are inaccurate or contradictory facts. We then compare the summary consistency before and after implementing the proposed methods. 

\section{Results}
\subsection{ROUGE Scores}
We report the changes of ROUGE scores\footnote{Absolute ROUGE scores are not shown due to confidentiality.} in Table \ref{table:ROUGE}. Results show that the models trained with our correction methods generally have improvements on the ROUGE scores compared to the original BART and T5 models. Higher scores imply that the summaries from the corrected models are better aligned with the target summaries. In addition, using $\mathcal{L}_{CC}$ as the loss function turns out to produce the highest ROUGE scores for both BART and T5. Thus, for human evaluation, we will focus on the summaries produced by the models trained with $\mathcal{L}_{CC}$.

\begin{table}
\centering
\begin{tabular}{lcc}
\hline
\textbf{Model} & \textbf{Error Type} & \textbf{\% Corrected} \\
\hline
\multirow{2}{*}{BART} & WED & 63.6 \\
 & IPD & 50.0 \\
\hline
\multirow{2}{*}{T5} & WED & 46.7 \\
 & IPD & 42.1 \\
\hline
\end{tabular}
\caption{\label{table:corrected} Percentage of corrected WED and IPD errors for BART and T5. Comparisons are made between the ordinary models and the models trained with $\mathcal{L}_{CC}$.}
\end{table}

\begin{table}
\centering
\begin{tabular}{lcc}
\hline
\textbf{Model} & \textbf{\% Consis. to Inconsist.} \\
\hline
BART & 1.2 \\
T5 & 2.1 \\
\hline
\end{tabular}
\caption{\label{table:inconsis} Percentage of cases where the summaries from the ordinary models are factual consistent but become inconsistent after our methods are applied.}
\end{table}

\subsection{Human Evaluation and Analysis}
The human evaluation included 124 examples for BART and 600 examples for T5, all of which were randomly sampled from the test set. Table \ref{table:corrected} shows the effect of our approaches on correcting the two major factual errors. As the results show, a large portion of the WED and IPD errors are corrected. Over 63\% WED and 50\% IPD mistakes from ordinary BART are rectified. For T5, our methods are able to correct around 46\% WED and 42\% IPD errors. It implies our models perform more robustly on the cases that can potentially lead to WED and IPD.  

One remaining question is whether our approaches would degrade the originally faithful summaries. In Table \ref{table:inconsis}, we report the percentage of cases where the summaries from the ordinary models are consistent but become inconsistent after using our methods. We can see that most of the summaries remain consistent from our models. Furthermore, our analysis shows that the overall amounts of inconsistent summaries are reduced by 44.1\% for BART and 31.6\% for T5, which indicates the effectiveness of our methods. 

Table \ref{table:correct-examples} shows several input texts and summaries from the models before and after using our methods. In the first example, our model is able to pick up the correct entity from multiple entities in the source document, where the ordinary model fails. In the second example, the summary from the ordinary model contains contradicting description against the source document but our model captures the correct information.

\begin{table}
\centering
\small
\begin{tabularx}{\columnwidth}{X}
\hline
\textbf{Source}: (...) I bought this expensive TV that's supposed to have good screen and built-in wifi connection. But this one runs with lots of lagging, not as advertised on the website. (…) \\ 
\textbf{Original}: \textcolor{red}{Screen} runs with lots of lagging, not as advertised. \\ 
\textbf{After}: \textcolor{blue}{TV} runs with lots of lagging, not as advertised. \\
\hline
\textbf{Source}: (…) The packaging is heavily damaged and opened, though the product inside is not broken. The seller should be careful on the packaging next time (…) \\ 
\textbf{Original}: The packaging is heavily damaged and opened. Product is  \textcolor{red}{broken}. \\ 
\textbf{After}: The packaging is heavily damaged and opened. The product inside is \textcolor{blue}{not broken}.\\
\hline
\end{tabularx}
\caption{\label{table:correct-examples} Examples of error corrections using our methods. }
\end{table}

\section{Conclusion}
In conclusion, we study the error patterns in the customer-feedback summaries generated by BART and T5. We propose to augment the training data with artificially corrupted summaries and use contrastive learning methods to enhance the model faithfulness. Human analysis shows that significant portion of WED and IPD errors from BART and T5 are reduced. Because our methods do not involve modifying the model structure, they can also be applied to other abstractive summarization frameworks.

\bibliographystyle{acl_natbib}
\bibliography{acl2021}

\end{document}